\documentclass[sigconf,nonacm]{acmart}

\setcopyright{none}
\renewcommand\footnotetextcopyrightpermission[1]{}

\usepackage{subcaption}
\usepackage{multirow}
\usepackage{enumitem}
\usepackage{pifont}
\usepackage{booktabs}
\usepackage{multirow}
\AtBeginDocument{%
  }

\begin{document}

\title{DOME: Learning Transferable Domain Variables from Sparse Supervision for Test-Time Adaptation}

\author{Xiaoran Xu}
\authornote{Equal contribution.}
\email{xuxiaoran22@mails.ucas.ac.cn}
\affiliation{%
  \institution{MAIS, IACAS}
  \city{Beijing}
  \country{China}
}

\author{Yifan Xu}
\authornotemark[1]
\email{yifan.xu@nlpr.ia.ac.cn}
\affiliation{%
  \institution{MAIS, IACAS}
  \city{Beijing}
  \country{China}
}

\author{Yupeng Wu}
\authornotemark[1]
\email{wuyupeng2023@ia.ac.cn}
\affiliation{%
  \institution{MAIS, IACAS}
  \city{Beijing}
  \country{China}
}

\author{Xiaoshan Yang}
\authornote{Corresponding author.}
\email{xiaoshan.yang@nlpr.ia.ac.cn}
\affiliation{%
  \institution{MAIS, IACAS}
  \city{Beijing}
  \country{China}
}

\author{Changsheng Xu}
\email{changsheng.xu@nlpr.ia.ac.cn}
\affiliation{%
  \institution{MAIS, IACAS}
  \city{Beijing}
  \country{China}
}

\renewcommand{\shortauthors}{Xu et al.}

\begin{abstract}
  Test-time adaptation (TTA) aims to align a model to shifting test domains using only unlabeled streaming data. Most existing methods implicitly infer a single global domain distribution, ignoring the multidimensional and sample-specific nature of real-world domain shifts, leading to fragile adaptation. We propose DOME, an effective domain encoder that explicitly models each sample’s domain in a zero-shot manner. DOME leverages vision-language pretraining to extract dense, continuous representations, parameterizes domains as distributional variables, and introduces a momentum-updated sparse domain bank for disentangled supervision. By injecting these explicit domain cues into downstream models, even a basic entropy-minimization TTA strategy achieves state-of-the-art performance across ImageNet-C, ImageNet-R, and ImageNet-Sketch, outperforming complex TTA approaches. Our results demonstrate that robust adaptation stems not from intricate adaptation algorithms, but from explicit, structured domain representation.
\end{abstract}

\maketitle

\section{Introduction}
The visual world is rich with diverse, continuous domains, such as climate change~\cite{hendrycks2019benchmarking}, image corruptions~\cite{hendrycks2021many}, and style transfer~\cite{jing2019neural}. These domain shifts make it challenging for models to generalize in real world. To address this, 
Test-Time Adaptation~(TTA)~\cite{wang2020tent,niu2023towards,wang2022continual} has emerged as a promising paradigm, which adapts the model to the target domain on the fly at inference with incoming, unlabeled test data. This direction is gaining increasing attention and prompting the development of more robust models.

Since the domain information is concealed within the visual input, which is not directly accessible, TTA methods typically seek the global statistical distribution as target domain during online optimization. 
Early TTA methods~\cite{wang2020tent,gong2022note,niu2022efficient} mitigate distribution shifts by adapting activation statistics and updating the affine parameters of normalization layers through lightweight self-supervised training, such as self-entropy~\cite{wang2020tent,niu2022efficient,wang2022continual}. 
Another line of research seeks to utilize the predictions of the model itself as pseudo-labels and re-train the model, known as the self-training strategy~\cite{chen2022contrastive, li2021free, ma2024improved, niu2022efficient}. 
Despite various approaches have been proposed, most works follow the ``implicit search'' paradigm, which treats the domain as a latent variable and searches for \textit{one single} distribution to represent all target data. However, domain representation is multidimensional. For example, an oil painting of birds can be categorized as either the artwork domain or the animal domain. This makes the implicit search fragile. First, the multidimensional, per-sample domain representation conflicts with the objective of learning a single, unified target-domain representation, which leads to unstable optimization and increases the risk of getting trapped in local minima. Second, one single domain representation for all target data eliminates sample-specific domain cues, causing information loss. Therefore previous TTA methods require carefully engineered heuristics to address these challenges.

\begin{figure}[t]
  \centering
    \centering
    \includegraphics[width=\linewidth]{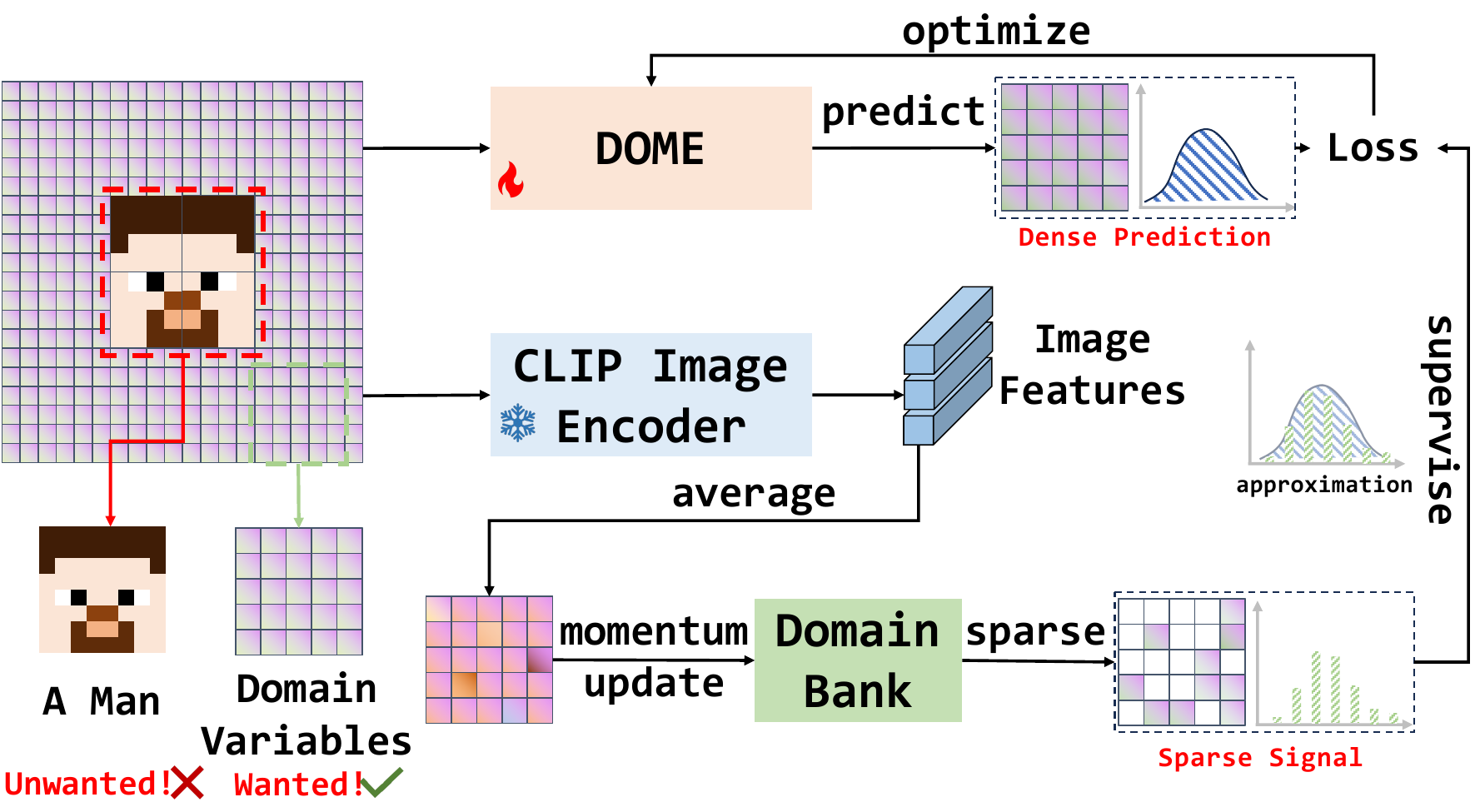}
    \caption{A simple illustration of DOME. DOME is supervised by a sparse momentum bank, which enables it to extract transferable domain representations instead of semantic information.
    }
    \label{fig:illustrate}
\end{figure}

This paper poses a simple question: \textit{can we explicitly model each sample’s domain instead of implicit search?}
With an explicit domain representation, test-time adaptation becomes data-dependent, allowing the inference process to be customized to each sample’s multidimensional domain rather than optimized for a global distribution. Moreover, explicitly providing domain information simplifies the task and allows the model to focus on semantic information.

To answer the above question,
we develop a domain encoder that can generate domain representations for any downstream samples in a zero-shot manner. Surprisingly, we found that even using self-entropy\cite{wang2020tent,niu2023towards}, a very naive TTA method, can achieve state-of-the-art performance on TTA benchmarks with explicit domain injection. However, it is nontrivial to obtain a domain encoder. 
The challenges and our insights are as follows.

\textbf{Dense domain encoding with vision-language pretraining}.
Domain representation is inherently continuous and multidimensional, whereas existing datasets~\cite{hendrycks2019benchmarking, peng2019moment, li2017deeper, sakaridis2021acdc, hendrycks2021many, wang2019learning, sakaridis2018semantic, hu2019depth, lin2014microsoft, cordts2016cityscapes,mao2023coco,richter2016playing} typically provide only discrete, single-domain labels. This mismatch makes it difficult to derive a domain representation from limited and discrete domain annotations from scratch.
We claim that pretrained vision-language models~\cite{radford2021learning} are well suited to address this gap: 1) Large-scale pretraining yields continuous, transferable representations. This enables the model to potentially learn a dense, continuous domain representation after light-weight fine-tuning~\cite{llava}. 2) Language serves as a scaffold for visual learning~\cite{radford2021learning,zhou2022conditional}. For example, explicitly instructing the model to extract the ``snowy'' domain largely reduces the chance it gets lost in the multidimensional domain space.


\textbf{Momentum sparse bank as supervision.}
Domain cues are entangled with semantics in images, and there is no direct supervision for domain alone. Several works~\cite{yuan2023robust, chen2021style, peng2019moment} maintain a feature bank to store dense features as domain signals using momentum updating strategies. However, since dense features remain entangled with semantic information, the resulting domain representations can only be applied to specific datasets.
Our key insight is that sparsifying output features can encourage disentanglement~\cite{cunningham2023sparse}, but it may hurt generalization and disrupt representation continuity when used directly for inference~\cite{petrini2022learning}.
Instead, we can use the sparse representation only as supervision signals. We construct a domain bank and update it with momentum sparse representation per sample. This reinforces shared domain information while suppressing sample-specific semantics.

Combining these insights, we propose a DOMain Encoder, DOME, 
capable of generating domain representations for any downstream sample in a zero-shot manner. Fig.~\ref{fig:illustrate} shows a simple explanation of DOME. DOME is learned by a Sparse-to-Dense (S2D) learning framework, which features: 1) \textbf{Dense domain encoding}: DOME is finetuned based on a CLIP~\cite{radford2021learning} image encoder to extract dense, continuous representations, parameterizing the final representation as means and variances for domain representation. We use partially masked domain text as guidance during training to avoid ambiguity in the multidimensional domain space. 2) \textbf{Sparse domain supervision}: We maintain a domain bank storing domain-related sparse features with a momentum updating strategy. We use this bank to supervise the main model. 3) \textbf{Explicit domain injection}. For downstream tasks, we use an MLP adapter to connect the frozen pretrained DOME with a downstream ViT, tuning only the adapter and normalization layers of the downstream ViT with lightweight self-entropy TTA objective.

In summary, the main contributions of this work are:
\begin{itemize}
    \item We propose a domain encoder, DOME, which can generate domain representation for any downstream samples in a zero-shot manner, transforming Test-Time domain Adaptation (TTA) from implicit domain search to explicit domain injection. To our best knowledge, DOME is the first general-purpose domain-specific feature extractor in the TTA literature.
    \item To train DOME, we propose a Sparse-to-Dense (S2D) learning framework featuring two key components:  
    (1) Sparse-Label Domain Disentanglement, constructing explicit supervision via a sparse momentum domain bank;
    (2) Dense Domain Representations Learning, extrapolating sparse annotations into a continuous domain representation space for zero-shot generalization.
    \item Experiments across multiple backbones and TTA benchmarks on both base and novel domains show that, with explicit domain injection, even a naive entropy-based TTA method can achieve state-of-the-art performance.
\end{itemize}

\section{Related Work}
\label{sec:formatting}

\subsection{Test-time Adaptation}

Test-time adaptation (TTA) has become a key paradigm for handling distribution shifts between training and test data, especially in real-world deployments where test conditions differ from the training environment~\cite{liang2025comprehensive,xiao2024beyond}. Unlike traditional domain adaptation, TTA adapts models at inference using test samples without class labels~\cite{wang2020tent}, making it well-suited for dynamic or unseen domains.

Existing TTA approaches primarily follow three paradigms: (i) Entropy-minimization, which refines model parameters by reducing prediction uncertainty~\cite{wang2020tent,lee2024entropy}; (ii) Prototype-based strategies, which maintain momentum-updated anchors to guide feature alignment~\cite{wang2022continual,niu2023towards}; and (iii) Self-supervised consistency, which enforces invariance across augmented views of test inputs~\cite{niu2024test}.

Despite their success, most existing TTA methods either operate without explicit domain modeling or infer domain characteristics solely from potentially noisy test-time predictions. This limits their ability to generalize under domain gaps and visual styles.




\subsection{Domain Representation Learning}
A key challenge in domain adaptation is disentangling domain-specific factors from domain-invariant semantics, as early methods often conflate these signals, harming generalization under distribution shift~\cite{ganin2016domain,ben2010theory}.  
Sparse Autoencoders (SAEs) promote disentanglement via sparse latent activations, producing interpretable features aligned with generative factors~\cite{ng2011sparse,makhzani2013k,cunningham2023sparse,gao2024scaling}, but their use for isolating domain-specific directions in TTA is limited.  
Feature-wise Linear Modulation (FiLM) injects external context through affine feature transformations~\cite{perez2018film}, enabling rapid adaptation without updating backbone weights~\cite{li2021universal,li2018domain,long2018conditional}, yet TTA rarely conditions on explicit, disentangled domain representations.  

Motivated by these advances, we integrate sparse domain encoding with FiLM-style modulation to construct a general-purpose domain-specific feature extractor, a functionality that existing TTA frameworks do not provide.

\begin{figure*}[t]
  \centering
  \begin{subfigure}{0.52\linewidth}
    \centering
    \includegraphics[width=\linewidth]{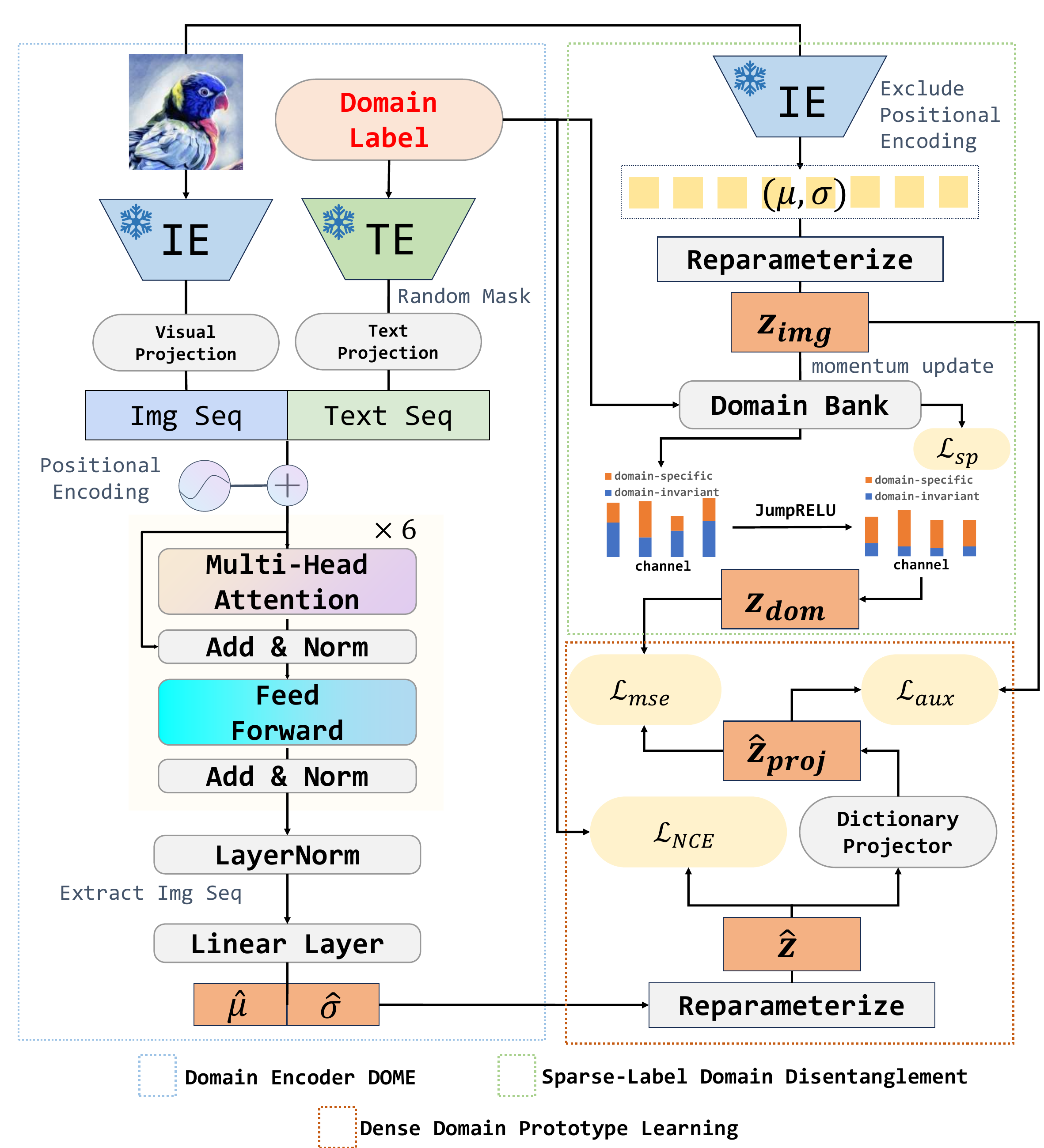}
    \caption{The DOME learns structured and transferable domain representations via Sparse-Label Domain Disentanglement and Dense Domain Representations Learning.}
    \label{fig:short-a}
  \end{subfigure}
  \hfill
  \begin{subfigure}{0.46\linewidth}
    \centering
    \includegraphics[width=\linewidth]{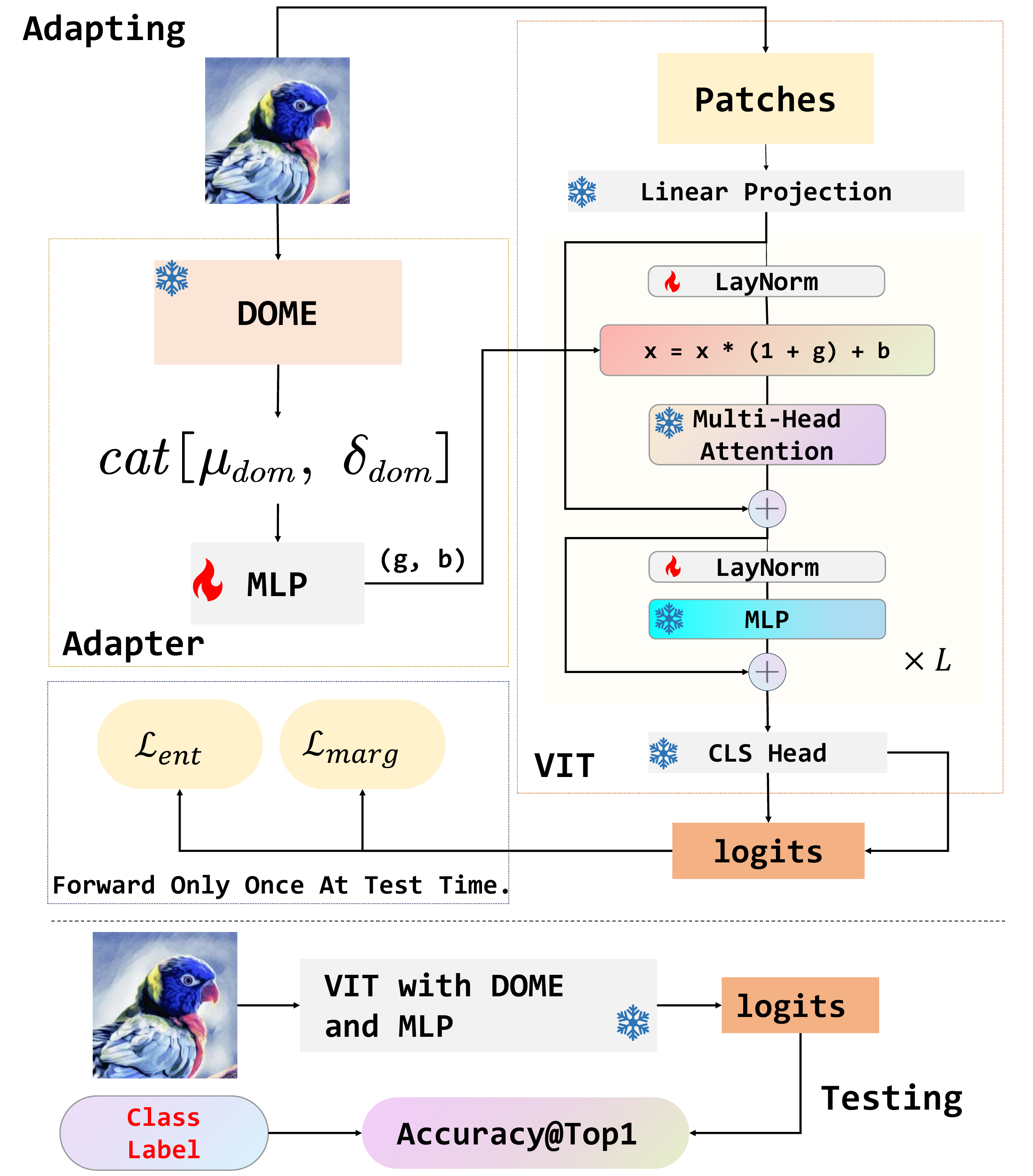}
    \caption{At test time, the domain statistics encoded by DOME are used to adapt the frozen VIT classifier via an adapter module, enabling robust recognition under distribution shifts.}
    \label{fig:short-b}
  \end{subfigure}
  \caption{Overview of our Sparse-to-Dense Learning Framework to train DOME and its integration with the VIT-based test-time adaptation.}
  \label{fig:short}
\end{figure*}

\section{Methodology}
\subsection{Overview}

To explicitly extract domain-specific features while suppressing category semantics and domain-invariant components, we propose the Sparse-to-Dense (S2D) framework for training the domain encoder DOME, shown in Fig.~\ref{fig:short-a}.  
DOME is trained on multiple domain datasets with domain labels but no category supervision, providing sparse yet explicit domain annotations for learning.

S2D consists of two synergistic components: (1) Sparse-Label Domain Disentanglement (SDD), isolating domain-specific factors from category semantics; and (2) Dense Domain Representation Learning (DDRL), extrapolating sparse annotations into a continuous domain space for sample-specific modeling and generalization to unseen domains.  

At deployment, DOME captures domain variables for Test-Time Adaptation (TTA), enabling adaptation to unseen target domains without class labels. Integrated with a lightweight TTA module, illustrated in Fig.~\ref{fig:short-b}, our method achieves state-of-the-art performance across benchmarks.  
Detailed descriptions follow.

\subsection{Sparse-to-Dense Learning Framework}
\noindent \textbf{Domain Encoder (DOME).}
The architecture of the domain encoder DOME is shown in Fig.~\ref{fig:short-a}. DOME captures domain-specific stylistic characteristics by fusing image and text features in a shared space, initialized from frozen CLIP's pre-trained image and text encoders. An input image is encoded into image tokens $f_{\text{img}}$, and its domain label into text tokens $f_{\text{txt}}$.
The text tokens provide a semantic prior that guides the extraction of domain variables, reducing ambiguity in the multidimensional domain representation. However, downstream samples typically lack domain labels. To mitigate the gap between DOME pretraining and downstream application, we randomly zero-mask the text embeddings with a probability of 20\%, yielding the partially masked domain text $\hat{f}_{\text{txt}}$.
These are then processed through six additional transformer layers to get the final fused image representation $f^{\prime}_{\text{img}}$, as:
\begin{equation}
\begin{aligned}
&f^{\prime}_{\text{img}}, f^{\prime}_{\text{txt}} = \text{Transformer}( [f_{\text{img}}; \hat{f}_{\text{txt}}]).
\end{aligned}
\end{equation}
During downstream implementation, we can simply use a zero embedding to replace the domain text $\hat{f}_{\text{txt}}$, eliminating the need for actual domain labels.

Inspired by the role of channel-wise statistics in style modeling~\cite{huang2017arbitrary,nam2018batch,ulyanov2016instance,zhou2021domain}, we represent domain-specific characteristics as predicted channel-wise mean $\hat{\mu}$ and standard deviation $\hat{\sigma}$. These are obtained by mapping $f^{\prime}_{\text{img}}$ through LayerNorm and a linear layer:
\begin{equation}
(\hat{\mu}, \hat{\sigma}) = \text{LinearLayer}(\text{Norm}(f^{\prime}_{\text{img}})).
\end{equation}
This compact statistical representation effectively captures the input image’s domain signature, enabling feature disentanglement and test-time adaptation.

\noindent \textbf{Sparse-Label Domain Disentanglement.}
In the absence of dense domain labels, extracting domain-specific information from limited supervision remains challenging. To address this, we propose \textit{Sparse-Label Domain Disentanglement} (SDD), which learns representations from the CLIP vision encoder that are discriminative across domains yet invariant to object categories, illustrated in Fig.~\ref{fig:short-a}.

We extract intermediate tokens from the CLIP vision encoder and remove their positional encodings to suppress category-related cues, encouraging focus on domain-specific appearance. From these position-free tokens, we compute channel-wise mean $\mu$ and standard deviation $\delta$ across spatial locations, yielding a compact global statistic $(\mu, \delta)$ as an initial content-driven proxy for domain style.

This statistic is reparameterized into a latent vector $\mathbf{z}_{\text{img}} \in \mathbb{R}^{d_{\text{hidden}}}$, representing the image’s raw stylistic signature. 
\begin{equation}
\mathbf{z}_{\text{img}} = \boldsymbol{\mu}_{\text{img}} + \boldsymbol{\sigma}_{\text{img}} \cdot \boldsymbol{\epsilon}, \qquad 
\boldsymbol{\epsilon} \sim \mathcal{N}(0, I).
\end{equation}





However, $\mathbf{z}_{\text{img}}$ often entangles domain factors with instance-level variations; using it directly to update the domain bank $\mathcal{B} = \{\mathbf{p}_{y_k} \mid k = 1, 2, \dots, K\}$, where ${y_k}$ is the $k$-th domain label and $\mathbf{p}_{y_k}$ is the learnable prototype for ${y_k}$, would propagate noise.

To mitigate this, we introduce a lightweight {Dictionary Decoder}, a linear mapping that lifts $\mathbf{z}_{\text{img}}$ into a high-dimensional, overcomplete domain space, enabling sparse and robust domain representations.
\begin{equation}
    \mathbf{d}_k = \mathbf{z}_{\text{img}} \mathbf{w}_{\text{dec}} + \mathbf{b}_{\text{dec}},
\end{equation}

Here, $\mathbf{w}_{\text{dec}} \in \mathbb{R}^{d_{\text{hidden}} \times d_{\text{in}}}$ with $d_{\text{in}} \gg d_{\text{hidden}}$, and its rows $\{\mathbf{p}_j^\top\}_{j=1}^{d_{\text{hidden}}}$ define learnable atoms in $\mathbb{R}^{d_{\text{in}}}$, each representing a canonical domain attribute. By lifting representations into this overcomplete space, the model activates only a sparse subset of interpretable features while suppressing irrelevant variations.

The resulting projections $\mathbf{d}_k$ update the domain bank via momentum update, where $\mathbf{d}_k$ is applied to the prototype $\mathbf{p}_{y_k}$:
\begin{equation}
    \mathbf{p}_{y_k} = m \cdot \mathbf{p}_{y_k} + (1 - m) \cdot \mathbf{d}_k,
\end{equation}

where the hyperparameter $m$ is set to $0.9$.

To enhance domain-specificity and disentangle domain-level features from variations, we apply $\ell_1$ sparsity regularization to the prototypes:
\begin{equation}
    \mathcal{L}_{sp} = \sum_{k=1}^{n_{\text{dom}}} \|\mathbf{m}_k\|_1,
\end{equation}
where $\mathbf{m}_k$ is the prototype for the $k$-th domain, $n_{\text{dom}}$ is the total numbers of domains. 
This sparsity encourages the domain bank to retain only consistent, recurring signals shared across instances within the same domain for training, while smoothing out noisy or irrelevant features. Combined with momentum updating, this ensures that the shared, domain-relevant features are reinforced, whereas instance-specific signals are gradually suppressed.

Finally, when using bank prototypes to supervise DOME, we apply \text{JumpReLU}~\cite{rajamanoharan2024jumping} to suppress residual weak activations. This element-wise sparsification on the retrieved prototype $\mathbf{m}_k$ yields a clean domain-specific representation $\mathbf{z}_{\text{dom}}$:
\begin{equation}
    \mathbf{z}_{\text{dom}}^{(i)} = \text{JumpReLU}(\mathbf{m}_k^{(i)}) = 
    \begin{cases} 
        \mathbf{m}_k^{(i)} & \text{if } \mathbf{m}_k^{(i)} \ge \beta, \\
        0 & \text{otherwise},
    \end{cases}
\end{equation}
where $\beta$ is a learnable threshold.

\noindent \textbf{Dense Domain Representations Learning.}
To align DOME’s features with the clean domain signals from SDD, we introduce Dense Domain Representations Learning (DDRL). Dense features $\hat{\mathbf{z}}$ capture full, sample-specific domain information beyond sparse prototypes, enabling robust and discriminative representations that generalize to unseen domains. As shown in Fig.~\ref{fig:short-a}, DDRL projects $\hat{\mathbf{z}}$, reparameterized from $(\hat{\mu}, \hat{\sigma})$, into an overcomplete space via a Dictionary Projection Layer to obtain $\hat{\mathbf{z}}_{\text{proj}}$, and enforces consistency with the sparse, high-fidelity signal $\mathbf{z}_{\text{dom}}$ through three losses.
First, an MSE loss regresses $\hat{\mathbf{z}}_{\text{proj}}$ toward $\mathbf{z}_{\text{dom}}$:
\begin{equation}
    \mathcal{L}_{\text{MSE}} = \left\| \hat{\mathbf{z}}_{\text{proj}} - \mathbf{z}_{\text{dom}} \right\|_2^2.
\end{equation}


To preserve projection fidelity, we add an auxiliary loss aligning $\hat{\mathbf{z}}_{\text{proj}}$ with the raw domain statistic $\mathbf{z}_{\text{img}}$:
\begin{equation}
    \mathcal{L}_{\text{aux}} = \left\| \hat{\mathbf{z}}_{\text{proj}} - \mathbf{z}_{\text{img}} \right\|_2^2.
\end{equation}

To encourage intra-domain cohesion and inter-domain separation, we use a multi-positive InfoNCE loss:
\begin{equation}
    \mathcal{L}_{\text{NCE}} = -\frac{1}{N} \sum_{i=1}^{N} \log \frac{\sum_{j \in P_i} \exp\big( \cos(\hat{\mathbf{z}}_i, \hat{\mathbf{z}}_j) / \tau \big)}{\sum_{k \neq i} \exp\big( \cos(\hat{\mathbf{z}}_i, \hat{\mathbf{z}}_k) / \tau \big)},
\end{equation}
where $P_i$ indexes samples from the same domain as $i$, $\cos(\cdot,\cdot)$ is cosine similarity, and $\tau$ is a temperature parameter. This yields domain-compact and cross-domain-discriminative representations.

\noindent \textbf{Overall Objective.}
The full training objective of the Sparse-to-Dense framework is a weighted sum of all introduced losses:
\begin{equation}
    \mathcal{L}_{\text{S2D}} = \lambda_{sp} \mathcal{L}_{sp} + \lambda_{\text{MSE}} \mathcal{L}_{\text{MSE}} + \lambda_{\text{NCE}} \mathcal{L}_{\text{NCE}} + \lambda_{\text{aux}} \mathcal{L}_{\text{aux}}.
\end{equation}
This composite loss jointly encourages sparsity in the domain bank, fidelity in projection, and cohesion among same-domain samples, enabling the learning of structured and transferable domain representations.

\subsection{Test-Time Adaptation via DOME}
Our frozen DOME is integrated into an Online Test-Time Adaptation (TTA) framework, shown in Fig.~\ref{fig:short-b}, dynamically generating domain-conditional normalization parameters for the ViT backbone to bridge sparse training supervision and dense test-time generalization.

DOME processes the original test image $\text{Img}_{\text{ori}}$ and its weakly augmented version $\text{Img}_{\text{aug}}$, producing domain statistics $(\mu_{\text{dom}}, \delta_{\text{dom}})$, which are fed into a lightweight MLP Adapter that outputs affine parameters $(g, b)$. These are injected after LayerNorm in each ViT block, modulating features $\mathbf{x}$ as
\begin{equation}
    \mathbf{x} = \mathbf{x} \cdot (1 + g) + b,
\end{equation}
which enables instance-level modulation that adapts features to the input’s specific style.

During inference, only the Adapter and LayerNorm affine terms are updated via a single gradient step; DOME and other ViT weights remain frozen. Adaptation uses unsupervised TTA losses: conditional entropy $\mathcal{L}_{\text{ent}}$~\cite{wang2020tent,lee2024entropy} and marginal entropy $\mathcal{L}_{\text{marg}}$~\cite{niu2022efficient,niu2023towards}, combined as
$\mathcal{L}_{\text{TTA}} = \lambda_{\text{ent}} \mathcal{L}_{\text{ent}} + \lambda_{\text{marg}} \mathcal{L}_{\text{marg}}$, with
\begin{equation}
    \mathcal{L}_{\text{ent}} = - \frac{1}{N} \sum_{i=1}^{N} \sum_{c=1}^{C} p_{i,c} \log p_{i,c}, \quad
    p_{i,c} = p(y=c \mid x_i),
\end{equation}
\begin{equation}
    \mathcal{L}_{\text{marg}} = \sum_{c=1}^{C} \bar{p}_c \log \bar{p}_c, \quad
    \bar{p}_c = \frac{1}{N} \sum_{i=1}^{N} p(y=c \mid x_i),
\end{equation}
where the marginal loss encourages high-entropy predictions to prevent collapse.

No additional supervision or task-specific design is used; performance gains derive entirely from the high-quality domain representations learned by S2D.

\begin{table*}[htbp]
\centering
\caption{Top-1 accuracy (\%) of different methods in the online test-time adaptation setting on ImageNet-C (severity level 5), where models adapt in a single pass over sequentially arriving batches without access to future data. Baseline: ViT-Base without test-time adaptation. Ours: Baseline + $\mathcal{L}_{\text{ent}}$ + $\mathcal{L}_{\text{marg}}$ + DOME.}
\label{tab:imagenet_c}
\resizebox{\textwidth}{!}{
\begin{tabular}{l ccc cccc cccc cccc c}
\toprule
\multirow{2}{*}{\textbf{Method}} &
\multicolumn{3}{c}{\textbf{Noise}} &
\multicolumn{4}{c}{\textbf{Blur}} &
\multicolumn{4}{c}{\textbf{Weather}} &
\multicolumn{4}{c}{\textbf{Digital}} &
\multirow{2}{*}{\textbf{Avg.}} \\ 
\cmidrule(lr){2-4} \cmidrule(lr){5-8} \cmidrule(lr){9-12} \cmidrule(lr){13-16}
& Gauss & Shot & Impul. 
& Defoc. & Glass & Motion & Zoom 
& Snow & Frost & Fog & Bright. 
& Contr. & Elastic & Pixel & JPEG \\ 
\midrule
Baseline & 56.8 & 56.8 & 57.5 & 46.9 & 35.6 & 53.1 & 44.8 & 62.2 & 62.5 & 65.7 & 77.7 & 32.6 & 46.0 & 67.0 & 67.6 & 55.5 \\
TENT   & 60.3 & 61.6 & 61.8 & 59.2 & 56.5 & 63.5 & 59.2 & 54.3 & 64.5 &  2.3 & 79.1 & 67.4 & 61.5 & 72.5 & 70.6 & 59.6 \\
CoTTA  & 63.6 & 63.8 & 64.1 & 55.5 & 51.1 & 63.6 & 55.5 & 70.0 & 69.4 & 71.5 & 78.5 &  46.7 & 64.5 & 73.4 & 71.2 & 64.2 \\
SAR    & 59.2 & 60.5 & 60.7 & 57.5 & 55.6 & 61.8 & 57.6 & 65.9 & 63.5 & 69.1 & 78.7 & 45.7 & 62.4 & 71.9 & 70.3 & 62.7 \\
ActMAD & 61.3 & 62.8 & 63.2 & 55.9 & 55.7 & 62.7 & 61.7 & 70.8 & 68.8 & 73.5 & 80.8 & 62.3 & 67.8 & 74.8 & 73.0 & 66.3 \\
DeYO   & 59.8 & 61.5 & 61.1 & 57.4 & 59.0 & 64.5 & 61.9 & 69.1 & 66.7 & 69.5 & 78.9 & 65.3 & 69.6 & 74.0 & 72.5 & 66.4 \\
FOA    & 61.5 & 63.2 & 63.3 & 59.3 & 56.7 & 61.4 & 57.7 & 69.4 & 69.6 & 73.4 & 81.1 & 67.7 & 62.7 & 73.9 & 73.0 & 66.3\\ 
\hline
Ours    & 58.8 & 60.7 & 60.6 & 58.8 & 58.7 & 64.7 & 64.4 & 70.5 & 68.9 & 72.7 & 79.9 & 62.2 & 70.8 & 75.8 & 73.0 &  \textbf{66.7} \\ 
\bottomrule
\end{tabular}
}
\end{table*}

\begin{table}[htbp]
\centering
\caption{Top-1 accuracy (\%) of different methods in the online test-time adaptation setting on ImageNet-R/-Sketch.}\label{tab:robustness_results}
\label{tab:imagenet-r/-sketch}
\begin{tabular}{l ccc}
\hline
\textbf{Method} & \textbf{R} & \textbf{Sketch} & \textbf{Avg.} \\
\hline
Baseline & 59.2 & 44.9 & 52.1 \\
TENT & 63.9 & 49.1 & 56.5 \\
CoTTA & 63.5 & 50.0 & 56.8 \\
SAR  & 63.3 & 48.7 & 56.0\\ 
ActMAD & 60.2 & 46.2 & 53.2 \\
DeYO & \textbf{68.7} & 50.3 & 59.5 \\
FOA & 63.3 & 49.9 & 56.6 \\
\hline
Ours & \textbf{68.7} & \textbf{53.9} & \textbf{61.3} \\
\bottomrule
\end{tabular}
\end{table}


\begin{table}[htbp]
\caption{Online TTA ablation. ``+$\mathcal{L}_{\text{ent}}$/+$\mathcal{L}_{\text{marg}}$'' uses conditional/marginal entropy with diversity regularization. }
\label{tab:online_tta_ablation}
\centering
\begin{tabular}{l ccc}
\hline
\textbf{Method} & \textbf{R} & \textbf{Sketch} & \textbf{Avg.} \\
\hline
Baseline & 59.2 & 44.9 & 52.1 \\
+$\mathcal{L}_{\text{ent}}$+$\mathcal{L}_{\text{marg}}$ & 62.9 & 49.9 & 56.6 \\ \hline
+$\mathcal{L}_{\text{ent}}$+$\mathcal{L}_{\text{marg}}$+DOME & \textbf{68.7} & \textbf{53.9} & \textbf{61.3} \\
\bottomrule
\end{tabular}
\end{table}

\begin{table}[htbp]
\caption{Ablation study on the training of the DOME using CLIP-Base, evaluated under online TTA on ImageNet-R. }\centering
\fontsize{9}{11}\selectfont
\setlength{\tabcolsep}{1.6mm}
\begin{tabular}{c|c|c}
\hline
Text & Sparsity  & $\text{Accuracy}$  \\ \hline
\ding{55} & \ding{55}& 59.2 \\
\ding{51} & \ding{55}& 67.5 \\
\ding{55} & \ding{51}& 67.0 \\
\ding{51} & \ding{51}& 68.7 \\ \hline
\end{tabular}
\label{tab:ablation_study}
\end{table}

\begin{table*}[htbp]
\centering
\caption{Offline TTA ablation across vision backbones and ImageNet-R. DomStat: uses target-domain statistics from a frozen CLIP Image encoder; Ours: employs our DOME. \textit{Seen} domains were unlabeled during training; \textit{Unseen} domains excluded.}
\label{tab:combined_domain_generalization_no_resize}
\setlength{\tabcolsep}{2.5pt} 
\renewcommand{\arraystretch}{1.1} 
\begin{tabular}{l | ccc | ccc | ccc | ccc}
\toprule
\textbf{Domain} & \multicolumn{3}{c|}{\textbf{ViT-B/16}} & \multicolumn{3}{c|}{\textbf{ViT-L/16}} & \multicolumn{3}{c|}{\textbf{DINOv2-B}} & \multicolumn{3}{c}{\textbf{DINOv2-L}} \\
\cmidrule(lr){2-4} \cmidrule(lr){5-7} \cmidrule(lr){8-10} \cmidrule(lr){11-13}
& \textbf{Base} & \textbf{DomStat} & \textbf{Ours} & \textbf{Base} & \textbf{DomStat} & \textbf{Ours} & \textbf{Base} & \textbf{DomStat} & \textbf{Ours} & \textbf{Base} & \textbf{DomStat} & \textbf{Ours} \\
\midrule
\multicolumn{13}{c}{\textit{\textbf{Seen Domains}}} \\
\midrule
cartoon & 45.94 & 36.85 & \textbf{56.58} & 53.01 & 52.98 & \textbf{61.35} & 57.94 & 61.35 & \textbf{70.61} & 72.09 & 73.67 & \textbf{74.21} \\
sketch & 73.97 & 49.33 & \textbf{81.25} & {80.84} & 77.97 & \textbf{84.07} & 85.37 & 86.04 & \textbf{88.02} & 92.58 & \textbf{93.18} & 92.60 \\
painting & 75.27 & 73.40 & \textbf{81.23} & 79.22 & 78.62 & \textbf{82.15} & 80.18 & 83.49 & \textbf{86.98} & 87.97 & \textbf{88.99} & 88.89 \\
videogame & 64.91 & 56.89 & \textbf{69.48} &  {69.27} & 67.37 &\textbf{69.13} & 67.44 & 73.77 & \textbf{78.83} & 75.88 & \textbf{79.18} & 78.27 \\
tattoo & 37.29 & 40.42 & \textbf{53.42} & 40.47 & 50.95 & \textbf{55.16} & 56.24 & 70.88 & \textbf{74.52} & 74.06 & 77.09 & \textbf{77.91} \\
\midrule
\multicolumn{13}{c}{\textit{\textbf{Unseen Domains}}} \\
\midrule
art & 62.30 & 57.59 & \textbf{70.22} & 66.56 & 68.46 & \textbf{73.49} & 66.23 & 74.08 & \textbf{79.32} & 78.08 & {81.28} & \textbf{81.54} \\
deviantart & 70.83 & 63.11 & \textbf{75.22} & 75.40 & 72.68 & \textbf{78.36} & 74.48 & 77.53 & \textbf{82.02} & 82.94 & {83.50} & \textbf{81.92} \\
embroidery & 37.81 & 43.49 & \textbf{58.45} & 41.55 & 53.74 & \textbf{59.14} & 42.80 & 68.14 & \textbf{77.42} & 60.25 & \textbf{75.90} & 73.27 \\
graffiti & 37.30 & 42.19 & \textbf{55.05} & 42.40 & 54.62 & \textbf{59.19} & 41.87 & 59.72 & \textbf{70.99} & 60.89 & 72.16 & \textbf{73.22} \\
graphic & 60.09 & 54.04 & \textbf{66.77} & 65.68 & 64.91 & \textbf{70.65} & 66.30 & 73.45 & \textbf{76.09} & 76.24 & 78.26 & \textbf{79.35} \\
misc & 60.51 & 58.29 & \textbf{69.93} & 66.01 & 67.23 & \textbf{72.10} & 68.82 & 73.68 & \textbf{77.84} & 78.28 & 80.65 & \textbf{81.60} \\
origami & 41.27 & 47.45 & \textbf{51.27} & 44.00 & 59.27 & \textbf{63.27} & 35.27 & 53.09 & \textbf{58.55} & 42.55 & 47.64 & \textbf{71.45} \\
sculpture & 55.27 & 52.93 & \textbf{63.71} & 60.95 & {63.96} & \textbf{67.22} & 64.46 & 71.32 & \textbf{75.33} & 79.18 & 82.11 & \textbf{83.44} \\
sticker & 44.69 & 41.93 & \textbf{56.50} & 48.82 & {55.31} & \textbf{60.63} & 49.21 & 64.96 & \textbf{68.90} & 65.94 & 72.24 & \textbf{75.19} \\
toy & 53.08 & 52.28 & \textbf{61.48} & 56.45 & 56.06 & \textbf{57.97} & 51.42 & 64.73 & \textbf{65.39} & 61.61 & 67.11 & \textbf{68.10} \\
\midrule
\textbf{Average} & 59.23 & 52.81 & \textbf{68.34} & 64.34 & 65.79 & \textbf{70.84} & 66.97 & 73.46 & \textbf{78.09} & 77.88 & 80.64 & \textbf{81.28} \\
\bottomrule
\end{tabular}
\end{table*}

\section{Experiment}
\label{sec:experiment}
\subsection{Experimental Setup}
\label{sec:experiment_setup}
\noindent \textbf{Dataset.}  
We construct a large-scale, diverse dataset named {DomainSpecific}, comprising 41 domains collected from multiples public datasets (e.g., ImageNet, DomainNet, PACS, et al.)~\cite{hendrycks2019benchmarking, peng2019moment, li2017deeper, sakaridis2021acdc, hendrycks2021many, wang2019learning, sakaridis2018semantic, hu2019depth, lin2014microsoft, cordts2016cityscapes, mao2023coco, richter2016playing}, with over 1.2M images. It covers a wide range of domain shifts, including artistic styles and real-world variations.

Following the standard unsupervised test-time adaptation (TTA) protocols, we discard class labels and retain only domain identities for the auxiliary representation learning. The \textit{DomainSpecific} data is partitioned into two disjoint subsets for training and evaluation:

\begin{itemize}[leftmargin=*, labelsep=0.5em, itemsep=0.25em]
    \item \textbf{Pretraining Domains (25):} To construct a high-diversity source domain pool for DOME, we integrate data from three distinct sources: 
    (i) \textbf{15 synthetic corruptions} generated by subsampling 50,000 images from the ImageNet-1K training dataset and applying perturbation protocols (severity level 5) as described in \cite{hendrycks2019benchmarking} (e.g., \textit{glass blur}, \textit{shot noise}, \textit{jpeg compression}); 
    (ii) \textbf{5 artistic domains} directly sourced from ImageNet-R (e.g., \textit{cartoon}, \textit{sketch}, \textit{painting}); 
    (iii) \textbf{5 diverse environmental and stylistic variations} collected from auxiliary datasets. 
    
    \item \textbf{OOD Domains (16):} These serve as strictly held-out targets to evaluate zero-shot generalization. This set includes:
    (i) \textbf{10 unseen artistic stylizations} from ImageNet-R (e.g., \textit{deviantart}, \textit{graffiti});
    (ii) \textbf{4 unseen corruption} categories from ImageNet-C (\textit{spatter}, \textit{speckle noise}, \textit{gaussian blur}, and \textit{saturate};
    (iii) \textbf{2 unseen artistic stylizations} from DomainNet (e.g. \textit{Real}, \textit{Infograph}).
\end{itemize}
Detailed domain configurations are provided in \textbf{Appendix}.





\noindent \textbf{Implementation Details.}  
Our DOME model uses CLIP image features with a 6-layer Transformer (8 attention heads) and mean pooling over patch tokens. Training runs for 30 epochs with AdamW (learning rate $1\times10^{-4}$, batch size 256). Domain prototypes are updated via momentum update~($m=0.9$), and loss weights are set as $\lambda_{sp}=1.0$, $\lambda_{\text{MSE}}=1.0$, $\lambda_{\text{NCE}}=0.1$, $\lambda_{\text{aux}}=0.1$. DOME employs CLIP-B and CLIP-L with hidden sizes $d_{\text{hidden}}=768$ and $1024$, respectively, and input dimension $d_{\text{in}}=16384$.

For classification, following CoTTA~\cite{wang2022continual} and SAR~\cite{niu2023towards}, we use SGD with momentum 0.9 and single GPU batch size 64. Online TTA adapts the model with one forward-backward pass per batch, while Offline TTA runs once over the full unlabeled test set, followed by re-evaluation. We set the entropy loss weight to $\lambda_{\text{ent}} = 1.4$ and the margin loss weight to $\lambda_{\text{marg}} = 0.8$.

For comparison, DomStat is a simple baseline we introduce, which averages CLIP image features from the target domain to form a fixed representation, illustrating DOME's ability to capture domain-specific information.
All experiments are conducted on NVIDIA RTX 5090 GPU, primarily using pretrained ViT-B and VIT-L models from timm~\cite{rw2019timm}.

\subsection{Performance Against Competing Methods}
Following standard online test-time adaptation protocols, we evaluate our method sequentially on ImageNet-C, ImageNet-R, and ImageNet-Sketch. The model adapts via a single forward-backward pass per batch in a streaming setting and reports overall accuracy.

Our method achieves the highest average accuracy across those benchmarks, outperforming TENT~\cite{wang2020tent}, CoTTA~\cite{wang2022continual}, SAR~\cite{niu2023towards}, ActMAD~\cite{mirza2023actmad}, DeYO~\cite{lee2024entropy}, and FOA~\cite{niu2024test}.

\textbf{ImageNet-C:} 
As shown in Tab.~\ref{tab:imagenet_c}, DOME combined with entropy minimization achieves SOTA accuracy (66.7\%), outperforming complex baselines like DeYO (66.4\%) and FOA (66.3\%).
Crucially, our protocol ensures zero data leakage: while pre-training involves synthetic domains, there is no overlap in either image samples or specific corruption instances with the ImageNet-C benchmark.
As shown in Tab.~\ref{tab:imagenet_c}, DOME combined with basic entropy minimization yields strong results, outperforming complex ensemble and memory-based methods. We achieve top performance on \textit{Motion Blur} (64.7\%), \textit{Zoom Blur} (64.4\%), \textit{Elastic Transform} (70.8\%) and \textit{Pixelate} (75.8\%), also remain stable on the other domains.

\textbf{ImageNet-R:} In Tab.~\ref{tab:imagenet-r/-sketch}, our method reaches \textbf{68.7\%} top-1 accuracy, euqal to DeYO (68.7\%), effectively handling styles shifts.

\textbf{ImageNet-Sketch:} We set a new state of the art at \textbf{53.9\%}, surpassing DeYO (50.3\%), showing strong robustness when texture cues disappear.

These results show that DOME allows simple entropy minimization to adapt effectively across diverse domain shifts.

\subsection{Ablation Study}
\noindent \textbf{Ablation on online-TTA.}
In online TTA, our method uses only a simple entropy-minimization objective, conditional entropy $\mathcal{L}_{\text{ent}}$ and marginal entropy $\mathcal{L}_{\text{marg}}$, without auxiliary regularization or complex update rules. As shown in Tab.~\ref{tab:online_tta_ablation}, it surpasses $\mathcal{L}_{\text{ent}}+\mathcal{L}_{\text{marg}}$ on both ImageNet-R and ImageNet-Sketch, highlighting the importance of high-quality domain-aware features for effective adaptation.


\noindent \textbf{Ablation on OOD Generalizability.}
We evaluate DOME under an offline TTA setting across diverse artistic domains. As shown in Tab.~\ref{tab:combined_domain_generalization_no_resize}, DOME consistently outperforms the DomStat baseline, which adapts using CLIP-derived domain statistics, on both seen and unseen domains and across two pretrained backbones~\cite{dosovitskiy2020image, oquab2023dinov2}.

With ViT-based models, our approach improves accuracy from 52.81\% to \textbf{68.34\%} on ViT-B/16 and from 65.79\% to \textbf{68.35\%} on ViT-L/16. Similar gains are observed with DINOv2, achieving \textbf{78.09\%} (B) and \textbf{81.28\%} (L). Notably, DOME yields substantial improvements on unseen domains in DOME's pretraining, highlighting its robustness to continuous domain shifts.

\noindent \textbf{Ablation on DOME.}
To isolate the contribution of key design choices in the DOME, we consider two variants:
\begin{itemize}
    \item \textbf{Ours w/o text}: trains DOME without CLIP text embeddings, using only visual inputs;
    \item \textbf{Ours w/o sparsity}: trains DOME without sparsity operation like JumpReLU, retaining all prototypes' responses.
\end{itemize}
As reported in Tab.~\ref{tab:ablation_study}, both variants result in performance drops, confirming that (1) semantic guidance from CLIP text prompts and (2) sparse feature selection via JumpReLU are essential for effective domain-specific representation learning.

In summary, our ablation studies validate that the strong performance of our method stems from the design of the DOME: high-quality domain-aware features, enabled by text guidance and sparse prototype activation, are the key drivers of adaptation success.

\begin{figure*}[t]
    \centering
    \includegraphics[width=\linewidth]{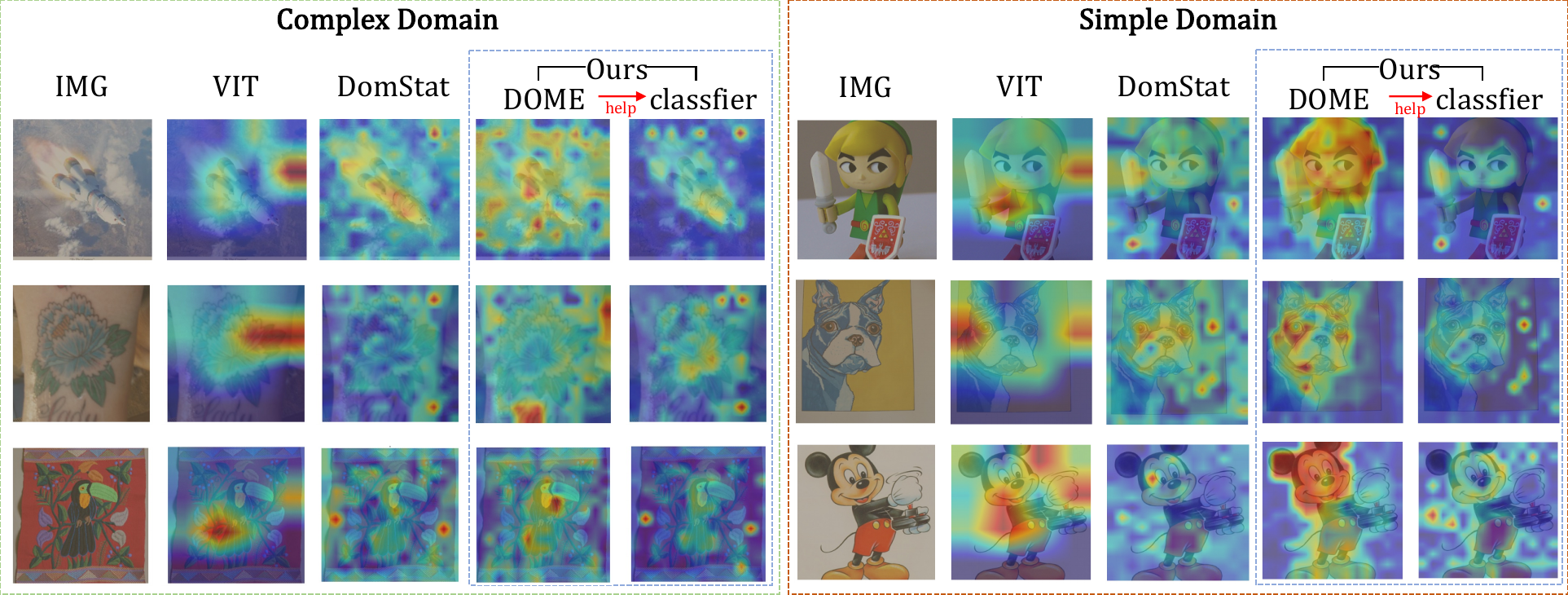}
    \caption{Attention maps~\cite{selvaraju2017grad} showing that our method produces sample-specific attention: it adaptively highlights domain-relevant context or object structure depending on the input, achieving better focus and semantic coverage than ViT/DomStat.}
    \label{fig:vis_analysis}
\end{figure*}

\begin{figure*}[t]
  \centering
  \begin{subfigure}{0.32\linewidth}
    \centering
    \includegraphics[width=\linewidth]{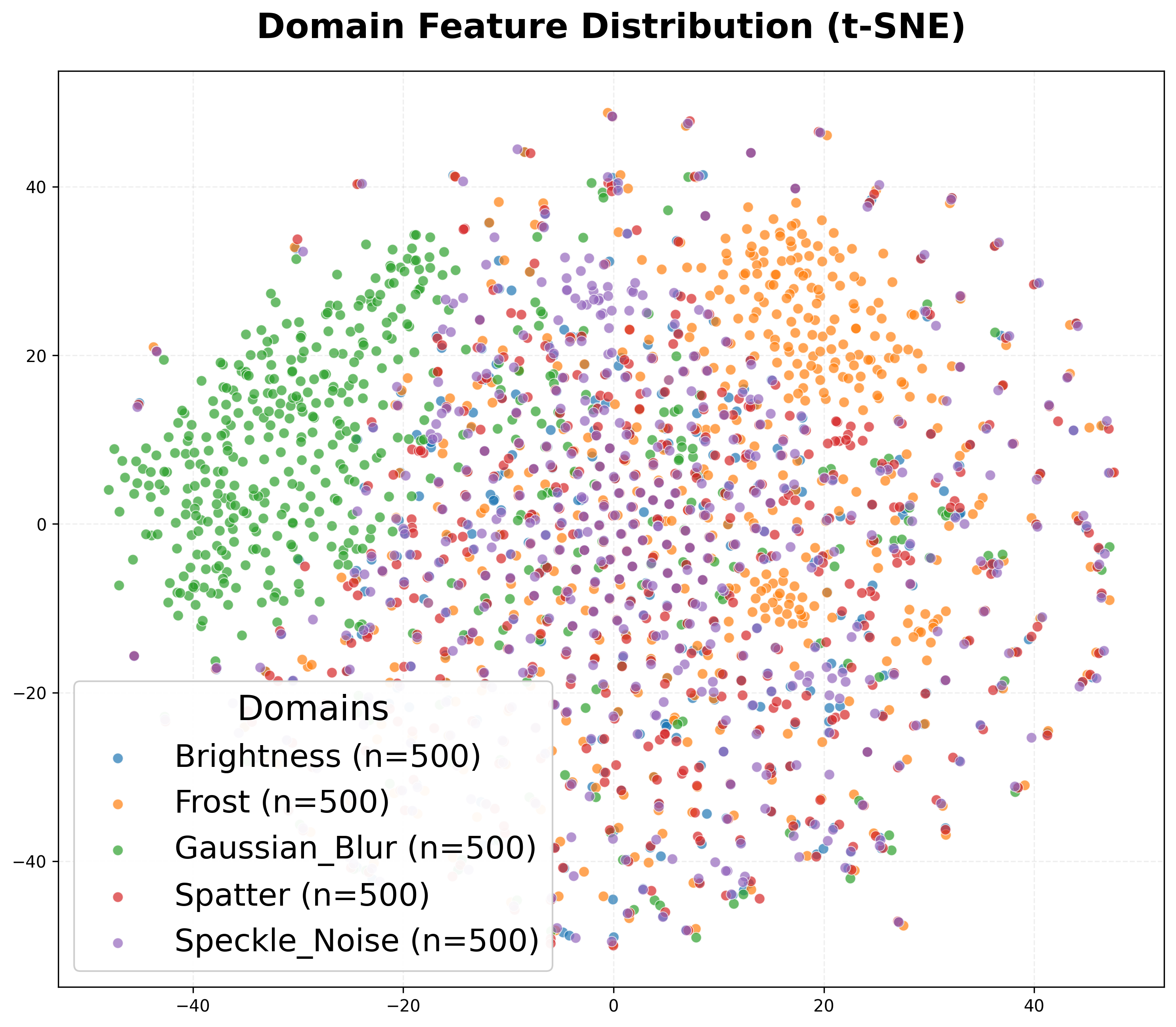}
    \caption{Features from the baseline (VIT-B/16) are highly entangled, showing poor domain separation.}
    \label{fig:tsne-vit}
  \end{subfigure}
  \hfill
  \centering
  \begin{subfigure}{0.32\linewidth}
    \centering
    \includegraphics[width=\linewidth]{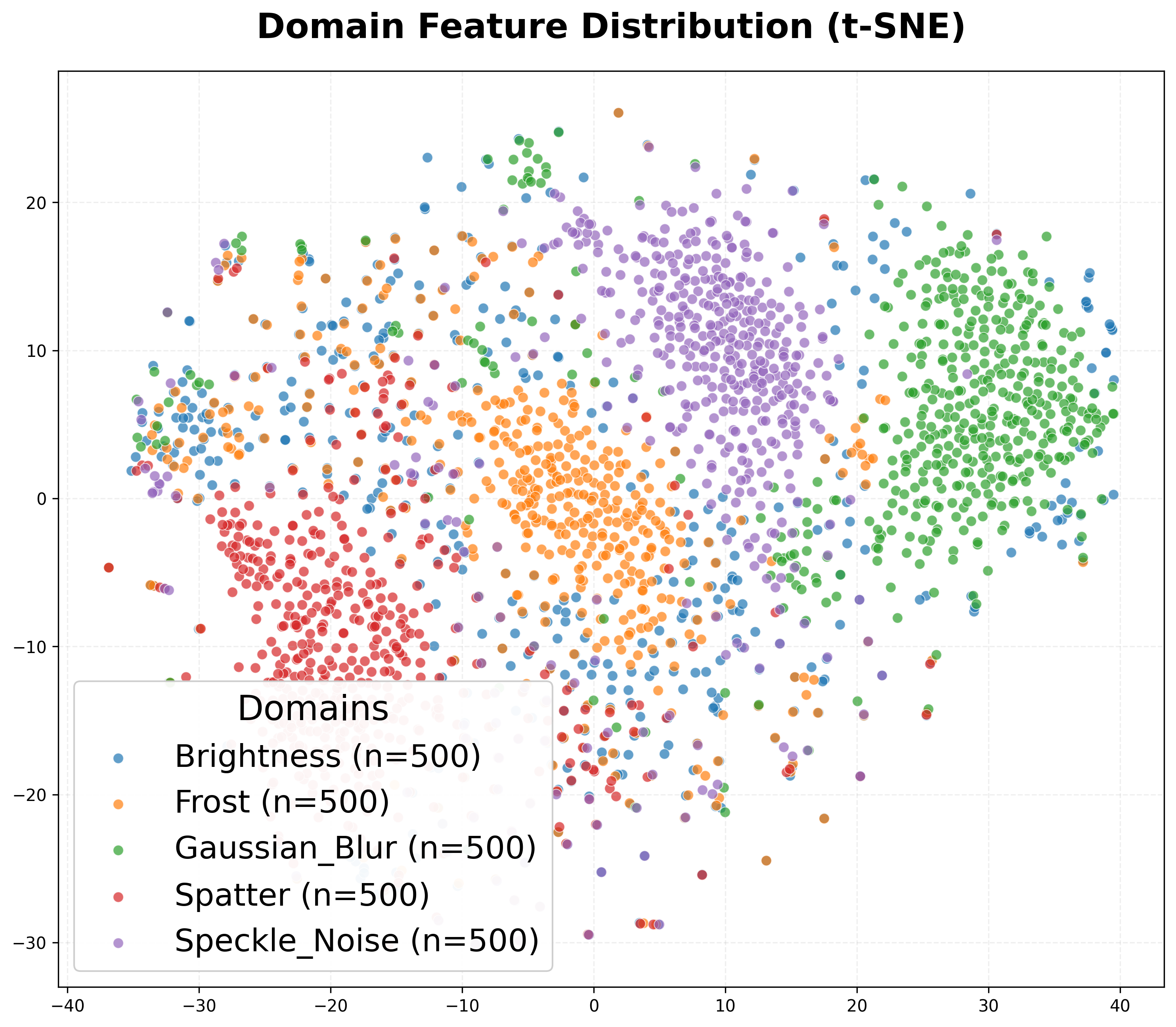}
    \caption{Features from the baseline (DomStat) are highly entangled, showing poor domain separation.}
    \label{fig:tsne-clip}
  \end{subfigure}
  \hfill
  \begin{subfigure}{0.32\linewidth}
    \centering
    \includegraphics[width=\linewidth]{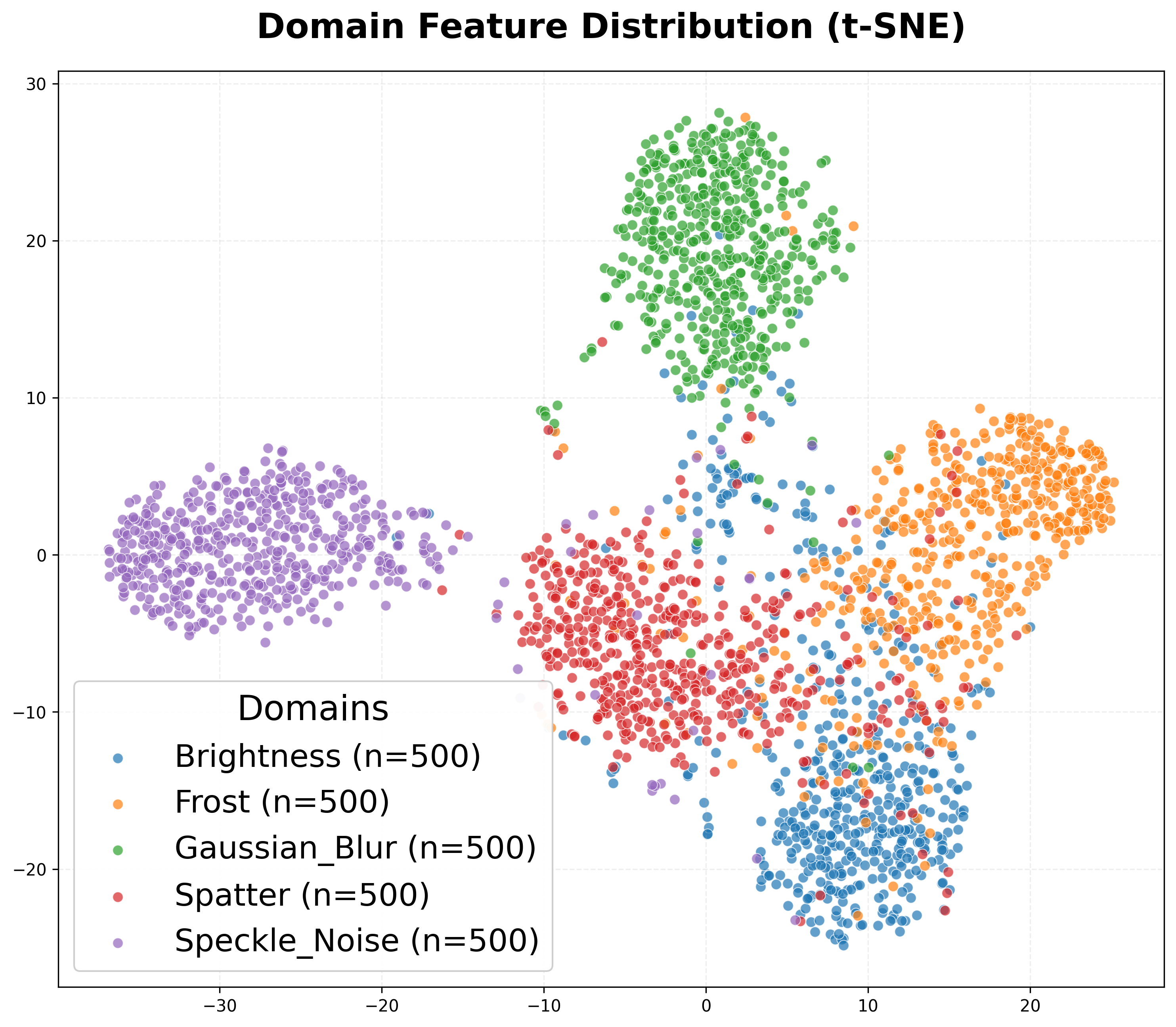}
    \caption{Features encoded with domain encoder DOME successfully form distinct, well-separated clusters.}
    \label{fig:tsne-ours}
  \end{subfigure}
    \caption{t-SNE visualization of feature distributions across five corruption domains from ImageNet-C: Brightness, Frost, Speckle Noise, Gaussian Blur, and Spatter, all at severity 5.}
\label{fig:tsne}
\end{figure*}

\subsection{Analysis and Visualization}
We provide qualitative and geometric insights into how DOME enables effective test-time adaptation by analyzing its attention behavior and learned feature structure.




\noindent \textbf{Attention Visualization.}
We examine DOME’s attention maps across diverse domains, shown in Fig.~\ref{fig:vis_analysis}, which reveal \textit{sample-specific} patterns that adapt to each input’s unique domain characteristics.  
For images with complex backgrounds and rich domain cues, DOME attends to both the foreground object and surrounding regions encoding domain-specific patterns, enabling accurate per-sample domain estimation and effective feature alignment. Conversely, when domain cues are sparse and mainly concentrated on the object itself, the attention map tightly focuses on the salient region, resulting in a heatmap that cleanly wraps around the entire object. Compared to ViT and DomStat, which may fragment attention or miss peripheral details, DOME’s holistic focus captures richer semantic features.  
This dual-mode behavior, contextual style when relevant and object integrity when necessary, demonstrates that DOME learns a flexible, sample-aware representation. 

\noindent \textbf{t-SNE Visualization.}
To analyze how DOME captures distribution shifts, we visualize feature distributions under five ImageNet-C corruptions: \textit{Brightness}, \textit{Frost}, \textit{Gaussian Blur}, \textit{Spatter}, and \textit{Speckle Noise}, each serving as a separate target domain during TTA.
As shown in Fig.~\ref{fig:tsne-vit} and Fig.~\ref{fig:tsne-clip}, ViT and DomStat produce highly entangled features across corruption types, revealing limited domain discrimination. In contrast, DOME (Fig.~\ref{fig:tsne-ours}) forms well-separated clusters aligned with domain structure, demonstrating its ability to learn discriminative domain representations.
These results show that DOME not only perceives domain-specific signals but also organizes its feature space according to test-time domain shifts, enabling more robust and interpretable adaptation.
These visualizations show that DOME separates domain-specific signals, yielding a structured feature space for robust and interpretable adaptation.

\section{Disscussion}
\noindent \textbf{DOME-Integrated Online TTA.}
To evaluate compatibility, we integrate DOME into representative online TTA baselines by equipping them with our pre-trained DOME. As shown in Tab.~\ref{tab:dome_plus_online_tta}, DOME consistently boosts all methods: TENT (+0.9\%), SAR (+1.8\%), CoTTA (+1.8\%), and DeYO (+0.3\%). This consistent improvement across entropy-minimization and stability-regularization schemes confirms that DOME’s explicit domain modeling provides critical robust cues that online self-supervision alone cannot capture.

\begin{table}[htbp]
    \centering
    \caption{Combining DOME with Existing Online TTA Methods. We report accuracy (\%) on ImageNet-R. ``+ DOME'' denotes modulating the model with our pre-trained DOME.}
    \label{tab:dome_plus_online_tta}
    \begin{tabular}{lcc}
        \toprule
        Method & w/o DOME & w/ DOME \\
        \midrule
        TENT~\cite{wang2020tent} & 63.9 & \textbf{64.8} \\
        SAR~\cite{niu2023towards} & 63.3 & \textbf{65.1} \\
        CoTTA~\cite{wang2022continual} &63.5 & \textbf{65.1} \\
        DeYO~\cite{lee2024entropy} & 68.7 & \textbf{69.0} \\
        \bottomrule
    \end{tabular}
\end{table}

\noindent \textbf{Computational Efficiency.}
DOME strikes an optimal trade-off between capacity and cost (Tab.~\ref{tab:compute_cost}): it achieves SOTA accuracy (66.7\%) using $4\times$ fewer trainable parameters than full-model adaptation (CoTTA: 21.53M vs. 86.57M), while maintaining real-time viability (17.56ms latency) with negligible overhead ($\approx 8.9$ms) over minimal-update methods (TENT). Although the dual-branch CLIP architecture increases total parameters (279.26M) and FLOPs (33.74G), this investment is justified by the unique ability to capture disentangled domain variables, a capability absent in implicit TTA baselines.
\begin{table}[htbp]
    \centering
    \caption{Comparison of Computational Overhead and Online TTA Performance on ImageNet-C. DOME offers an optimal balance between Trainable Parameters and expressivity.}
    \label{tab:compute_cost}
    \resizebox{\linewidth}{!}{
    \begin{tabular}{l|cc|cc|c}
        \toprule
        \multirow{2}{*}{\textbf{Method}} & \multicolumn{2}{c|}{\textbf{Complexity}} & \multicolumn{2}{c|}{\textbf{Parameters (M)}} & \textbf{Performance} \\
         & \textbf{FLOPs (G)} & \textbf{Latency (ms)} & \textbf{Trainable} & \textbf{Total} & \textbf{Avg. Acc (\%)} \\
        \midrule
        \textbf{TENT}~\cite{wang2020tent} & 16.87 & 8.66 $\pm 0.48$ & 0.04 & 86.57 & 59.6 \\
        \textbf{CoTTA}~\cite{wang2022continual} & 16.87 & 394.57 ± 29.27 & 86.57 & 86.57 & 63.5 \\
        \midrule
        \textbf{DOME+TTA} & 33.74 & 17.56 $\pm 0.25$ & 21.53 & 279.26 & 66.7 \\
        \bottomrule
    \end{tabular}
    }
\end{table}

\noindent \textbf{Generalization Beyond Natural Image Manifolds.}
To further assess the robustness of our approach under more severe distribution shifts, we extend our evaluation to DomainNet~\cite{peng2019moment}, which contains substantial cross-domain variations beyond natural image statistics.
Specifically, we fine-tune a ViT-Base model on the \textit{Real} domain and evaluate its generalization performance on the \textit{Infograph} domain, representing a challenging transfer from natural images to highly abstract visual representations.

As shown in Tab.~\ref{tab:domainnet_tta}, the standard fine-tuned ViT-Base achieves 24.69\% Top-1 accuracy. By incorporating DOME with offline test-time adaptation, the performance improves to 25.80\% Top-1 and 40.74\% Top-5 accuracy, yielding gains of +1.11\% and +3.46\%, respectively.
These consistent improvements under significant domain discrepancy suggest that DOME enhances the model's ability to capture domain-invariant features, thereby improving generalization beyond the ImageNet-style distribution.
\begin{table}[ht]
    \centering
    \caption{Generalization results on the DomainNet. Models are fintuned on the \textit{Real} domain and evaluated on the \textit{Infograph} domain, which are all not included in DOME's pretraining, to test robustness against non-natural imaging shifts.}
    \label{tab:domainnet_tta}
    \small
    \begin{tabular}{lcc}
        \toprule
        \multirow{2}{*}{\textbf{Method}} & \multicolumn{2}{c}{\textbf{Real $\to$ Infograph}} \\
        \cmidrule(lr){2-3}
        & \textbf{Top-1 Acc. (\%)} & \textbf{Top-5 Acc. (\%)} \\
        \midrule
        Full Fine-tuning & 24.69 & 37.28 \\
        Full Fine-tuning + \textbf{DOME} & \textbf{25.80} & \textbf{40.74} \\
        \bottomrule
    \end{tabular}
\end{table}

\section{Conclusion}
We propose DOME, a zero-shot domain encoder, which transforms test-time adaptation from an implicit domain search paradigm to an explict domain injection paradigm. By leveraging vision-language pretraining, representing domains as distributional variables, and using sparse momentum-based supervision, DOME successfully disentangles domain cues from semantics.  
Remarkably, DOME with simple entropy minimization achieves state-of-the-art performance on diverse unseen domains, showing that robust adaptation relies on explicit domain understanding.  
We hope DOME inspires more explicit domain modeling for adaptive vision in open-world settings.

\begin{acks}
To Robert, for the bagels and explaining CMYK and color spaces.
\end{acks}

\bibliographystyle{ACM-Reference-Format}
\bibliography{sample-base}

@String{Computer = "{IEEE} Computer" }

@String{Springer = "Springer-Verlag" }

@article{liang2025comprehensive,
  title={A comprehensive survey on test-time adaptation under distribution shifts},
  author={Liang, Jian and He, Ran and Tan, Tieniu},
  journal={International Journal of Computer Vision},
  volume={133},
  number={1},
  pages={31--64},
  year={2025},
  publisher={Springer}
}

@article{xiao2024beyond,
  title={Beyond model adaptation at test time: A survey},
  author={Xiao, Zehao and Snoek, Cees GM},
  journal={arXiv preprint arXiv:2411.03687},
  year={2024}
}

@article{wang2020tent,
  title={Tent: Fully test-time adaptation by entropy minimization},
  author={Wang, Dequan and Shelhamer, Evan and Liu, Shaoteng and Olshausen, Bruno and Darrell, Trevor},
  journal={arXiv preprint arXiv:2006.10726},
  year={2020}
}

@inproceedings{huang2017arbitrary,
  title={Arbitrary style transfer in real-time with adaptive instance normalization},
  author={Huang, Xun and Belongie, Serge},
  booktitle={Proceedings of the IEEE international conference on computer vision},
  pages={1501--1510},
  year={2017}
}

@article{ulyanov2016instance,
  title={Instance normalization: The missing ingredient for fast stylization},
  author={Ulyanov, Dmitry and Vedaldi, Andrea and Lempitsky, Victor},
  journal={arXiv preprint arXiv:1607.08022},
  year={2016}
}

@article{zhou2021domain,
  title={Domain generalization with mixstyle},
  author={Zhou, Kaiyang and Yang, Yongxin and Qiao, Yu and Xiang, Tao},
  journal={arXiv preprint arXiv:2104.02008},
  year={2021}
}

@article{nam2018batch,
  title={Batch-instance normalization for adaptively style-invariant neural networks},
  author={Nam, Hyeonseob and Kim, Hyo-Eun},
  journal={Advances in Neural Information Processing Systems},
  volume={31},
  year={2018}
}

@article{rajamanoharan2024jumping,
  title={Jumping ahead: Improving reconstruction fidelity with jumprelu sparse autoencoders},
  author={Rajamanoharan, Senthooran and Lieberum, Tom and Sonnerat, Nicolas and Conmy, Arthur and Varma, Vikrant and Kram{\'a}r, J{\'a}nos and Nanda, Neel},
  journal={arXiv preprint arXiv:2407.14435},
  year={2024}
}

@inproceedings{mirza2023actmad,
  title={Actmad: Activation matching to align distributions for test-time-training},
  author={Mirza, Muhammad Jehanzeb and Soneira, Pol Jan{\'e} and Lin, Wei and Kozinski, Mateusz and Possegger, Horst and Bischof, Horst},
  booktitle={Proceedings of the IEEE/CVF Conference on Computer Vision and Pattern Recognition},
  pages={24152--24161},
  year={2023}
}

@article{niu2023towards,
  title={Towards stable test-time adaptation in dynamic wild world},
  author={Niu, Shuaicheng and Wu, Jiaxiang and Zhang, Yifan and Wen, Zhiquan and Chen, Yaofo and Zhao, Peilin and Tan, Mingkui},
  journal={arXiv preprint arXiv:2302.12400},
  year={2023}
}

@inproceedings{wang2022continual,
  title={Continual test-time domain adaptation},
  author={Wang, Qin and Fink, Olga and Van Gool, Luc and Dai, Dengxin},
  booktitle={Proceedings of the IEEE/CVF Conference on Computer Vision and Pattern Recognition},
  pages={7201--7211},
  year={2022}
}

@inproceedings{richter2016playing,
title={Playing for data: Ground truth from computer games},
author={Richter, Stephan R and Vineet, Vibhav and Roth, Stefan and Koltun, Vladlen},
booktitle={European conference on computer vision},
pages={102--118},
year={2016},
organization={Springer}
}

@misc{rw2019timm,
  author = {Ross Wightman},
  title = {PyTorch Image Models},
  year = {2019},
  publisher = {GitHub},
  journal = {GitHub repository},
  doi = {10.5281/zenodo.4414861},
  howpublished = {\url{https://github.com/rwightman/pytorch-image-models}}
}

@article{lee2024entropy,
  title={Entropy is not enough for test-time adaptation: From the perspective of disentangled factors},
  author={Lee, Jonghyun and Jung, Dahuin and Lee, Saehyung and Park, Junsung and Shin, Juhyeon and Hwang, Uiwon and Yoon, Sungroh},
  journal={arXiv preprint arXiv:2403.07366},
  year={2024}
}

@article{niu2024test,
  title={Test-time model adaptation with only forward passes},
  author={Niu, Shuaicheng and Miao, Chunyan and Chen, Guohao and Wu, Pengcheng and Zhao, Peilin},
  journal={arXiv preprint arXiv:2404.01650},
  year={2024}
}

@inproceedings{mao2023coco,
  title={Coco-o: A benchmark for object detectors under natural distribution shifts},
  author={Mao, Xiaofeng and Chen, Yuefeng and Zhu, Yao and Chen, Da and Su, Hang and Zhang, Rong and Xue, Hui},
  booktitle={Proceedings of the IEEE/CVF International Conference on Computer Vision},
  pages={6339--6350},
  year={2023}
}

@article{hendrycks2019benchmarking,
  title={Benchmarking neural network robustness to common corruptions and perturbations},
  author={Hendrycks, Dan and Dietterich, Thomas},
  journal={arXiv preprint arXiv:1903.12261},
  year={2019},
  note={Includes ImageNet-C}
}

@inproceedings{hendrycks2021many,
  title={The many faces of robustness: A critical analysis of out-of-distribution generalization},
  author={Hendrycks, Dan and Basart, Steven and Mu, Norman and Kadavath, Saurav and Wang, Frank and Dorundo, Evan and Desai, Rahul and Zhu, Tyler and Parajuli, Samyak and Guo, Mike and others},
  booktitle={Proceedings of the IEEE/CVF international conference on computer vision},
  pages={8340--8349},
  year={2021}
}

@inproceedings{peng2019moment,
  title={Moment matching for multi-source domain adaptation},
  author={Peng, Xingchao and Bai, Qinxun and Xia, Xide and Huang, Zijun and Saenko, Kate and Wang, Bo},
  booktitle={Proceedings of the IEEE/CVF international conference on computer vision},
  pages={1406--1415},
  year={2019}
}

@inproceedings{li2017deeper,
  title={Deeper, broader and artier domain generalization},
  author={Li, Da and Yang, Yongxin and Song, Yi-Zhe and Hospedales, Timothy M},
  booktitle={Proceedings of the IEEE international conference on computer vision},
  pages={5542--5550},
  year={2017}
}

@inproceedings{sakaridis2021acdc,
  title={ACDC: The adverse conditions dataset with correspondences for semantic driving scene understanding},
  author={Sakaridis, Christos and Dai, Dengxin and Van Gool, Luc},
  booktitle={Proceedings of the IEEE/CVF international conference on computer vision},
  pages={10765--10775},
  year={2021}
}

@inproceedings{lin2014microsoft,
  title={Microsoft coco: Common objects in context},
  author={Lin, Tsung-Yi and Maire, Michael and Belongie, Serge and Hays, James and Perona, Pietro and Ramanan, Deva and Doll{\'a}r, Piotr and Zitnick, C Lawrence},
  booktitle={European conference on computer vision},
  pages={740--755},
  year={2014},
  organization={Springer}
}

@inproceedings{cordts2016cityscapes,
  title={The cityscapes dataset for semantic urban scene understanding},
  author={Cordts, Marius and Omran, Mohamed and Ramos, Sebastian and Rehfeld, Timo and Enzweiler, Markus and Benenson, Rodrigo and Franke, Uwe and Roth, Stefan and Schiele, Bernt},
  booktitle={Proceedings of the IEEE conference on computer vision and pattern recognition},
  pages={3213--3223},
  year={2016}
}

@article{sakaridis2018semantic,
  title={Semantic foggy scene understanding with synthetic data},
  author={Sakaridis, Christos and Dai, Dengxin and Van Gool, Luc},
  journal={International Journal of Computer Vision},
  volume={126},
  number={9},
  pages={973--992},
  year={2018},
  publisher={Springer}
}

@inproceedings{hu2019depth,
  title={Depth-attentional features for single-image rain removal},
  author={Hu, Xiaowei and Fu, Chi-Wing and Zhu, Lei and Heng, Pheng-Ann},
  booktitle={Proceedings of the IEEE/CVF Conference on computer vision and pattern recognition},
  pages={8022--8031},
  year={2019}
}

@article{wang2019learning,
  title={Learning robust global representations by penalizing local predictive power},
  author={Wang, Haohan and Ge, Songwei and Lipton, Zachary and Xing, Eric P},
  journal={Advances in neural information processing systems},
  volume={32},
  year={2019}
}

@article{makhzani2013k,
  title={K-sparse autoencoders},
  author={Makhzani, Alireza and Frey, Brendan},
  journal={arXiv preprint arXiv:1312.5663},
  year={2013}
}

@article{long2018conditional,
  title={Conditional adversarial domain adaptation},
  author={Long, Mingsheng and Cao, Zhangjie and Wang, Jianmin and Jordan, Michael I},
  journal={Advances in neural information processing systems},
  volume={31},
  year={2018}
}

@inproceedings{li2018domain,
  title={Domain generalization via conditional invariant representations},
  author={Li, Ya and Gong, Mingming and Tian, Xinmei and Liu, Tongliang and Tao, Dacheng},
  booktitle={Proceedings of the AAAI conference on artificial intelligence},
  volume={32},
  number={1},
  year={2018}
}

@inproceedings{li2021universal,
  title={Universal representation learning from multiple domains for few-shot classification},
  author={Li, Wei-Hong and Liu, Xialei and Bilen, Hakan},
  booktitle={Proceedings of the IEEE/CVF international conference on computer vision},
  pages={9526--9535},
  year={2021}
}

@inproceedings{perez2018film,
  title={Film: Visual reasoning with a general conditioning layer},
  author={Perez, Ethan and Strub, Florian and De Vries, Harm and Dumoulin, Vincent and Courville, Aaron},
  booktitle={Proceedings of the AAAI conference on artificial intelligence},
  volume={32},
  number={1},
  year={2018}
}

@article{gao2024scaling,
  title={Scaling and evaluating sparse autoencoders},
  author={Gao, Leo and la Tour, Tom Dupr{\'e} and Tillman, Henk and Goh, Gabriel and Troll, Rajan and Radford, Alec and Sutskever, Ilya and Leike, Jan and Wu, Jeffrey},
  journal={arXiv preprint arXiv:2406.04093},
  year={2024}
}

@article{ng2011sparse,
  title={Sparse autoencoder},
  author={Ng, Andrew and others},
  journal={CS294A Lecture notes},
  volume={72},
  number={2011},
  pages={1--19},
  year={2011}
}

@article{cunningham2023sparse,
  title={Sparse autoencoders find highly interpretable features in language models},
  author={Cunningham, Hoagy and Ewart, Aidan and Riggs, Logan and Huben, Robert and Sharkey, Lee},
  journal={arXiv preprint arXiv:2309.08600},
  year={2023}
}

@article{ben2010theory,
  title={A theory of learning from different domains},
  author={Ben-David, Shai and Blitzer, John and Crammer, Koby and Kulesza, Alex and Pereira, Fernando and Vaughan, Jennifer Wortman},
  journal={Machine learning},
  volume={79},
  number={1},
  pages={151--175},
  year={2010},
  publisher={Springer}
}

@article{ganin2016domain,
  title={Domain-adversarial training of neural networks},
  author={Ganin, Yaroslav and Ustinova, Evgeniya and Ajakan, Hana and Germain, Pascal and Larochelle, Hugo and Laviolette, Fran{\c{c}}ois and March, Mario and Lempitsky, Victor},
  journal={Journal of machine learning research},
  volume={17},
  number={59},
  pages={1--35},
  year={2016}
}

@inproceedings{niu2022efficient,
  title={Efficient test-time model adaptation without forgetting},
  author={Niu, Shuaicheng and Wu, Jiaxiang and Zhang, Yifan and Chen, Yaofo and Zheng, Shijian and Zhao, Peilin and Tan, Mingkui},
  booktitle={International conference on machine learning},
  pages={16888--16905},
  year={2022},
  organization={PMLR}
}

@inproceedings{selvaraju2017grad,
  title={Grad-cam: Visual explanations from deep networks via gradient-based localization},
  author={Selvaraju, Ramprasaath R and Cogswell, Michael and Das, Abhishek and Vedantam, Ramakrishna and Parikh, Devi and Batra, Dhruv},
  booktitle={Proceedings of the IEEE international conference on computer vision},
  pages={618--626},
  year={2017}
}

@article{oquab2023dinov2,
  title={Dinov2: Learning robust visual features without supervision},
  author={Oquab, Maxime and Darcet, Timoth{\'e}e and Moutakanni, Th{\'e}o and Vo, Huy and Szafraniec, Marc and Khalidov, Vasil and Fernandez, Pierre and Haziza, Daniel and Massa, Francisco and El-Nouby, Alaaeldin and others},
  journal={arXiv preprint arXiv:2304.07193},
  year={2023}
}

@article{dosovitskiy2020image,
  title={An image is worth 16x16 words: Transformers for image recognition at scale},
  author={Dosovitskiy, Alexey},
  journal={arXiv preprint arXiv:2010.11929},
  year={2020}
}

@inproceedings{yuan2023robust,
  title={Robust test-time adaptation in dynamic scenarios},
  author={Yuan, Longhui and Xie, Binhui and Li, Shuang},
  booktitle={Proceedings of the IEEE/CVF Conference on Computer Vision and Pattern Recognition},
  pages={15922--15932},
  year={2023}
}

@inproceedings{chen2021style,
  title={A style and semantic memory mechanism for domain generalization},
  author={Chen, Yang and Wang, Yu and Pan, Yingwei and Yao, Ting and Tian, Xinmei and Mei, Tao},
  booktitle={Proceedings of the IEEE/CVF International Conference on Computer Vision},
  pages={9164--9173},
  year={2021}
}

@article{petrini2022learning,
  title={Learning sparse features can lead to overfitting in neural networks},
  author={Petrini, Leonardo and Cagnetta, Francesco and Vanden-Eijnden, Eric and Wyart, Matthieu},
  journal={Advances in Neural Information Processing Systems},
  volume={35},
  pages={9403--9416},
  year={2022}
}

@inproceedings{radford2021learning,
  title={Learning transferable visual models from natural language supervision},
  author={Radford, Alec and Kim, Jong Wook and Hallacy, Chris and Ramesh, Aditya and Goh, Gabriel and Agarwal, Sandhini and Sastry, Girish and Askell, Amanda and Mishkin, Pamela and Clark, Jack and others},
  booktitle={International conference on machine learning},
  pages={8748--8763},
  year={2021},
  organization={PmLR}
}

@inproceedings{chen2022contrastive,
  title={Contrastive test-time adaptation},
  author={Chen, Dian and Wang, Dequan and Darrell, Trevor and Ebrahimi, Sayna},
  booktitle={Proceedings of the IEEE/CVF Conference on Computer Vision and Pattern Recognition},
  pages={295--305},
  year={2022}
}

@inproceedings{li2021free,
  title={A free lunch for unsupervised domain adaptive object detection without source data},
  author={Li, Xianfeng and Chen, Weijie and Xie, Di and Yang, Shicai and Yuan, Peng and Pu, Shiliang and Zhuang, Yueting},
  booktitle={Proceedings of the AAAI Conference on Artificial Intelligence},
  volume={35},
  number={10},
  pages={8474--8481},
  year={2021}
}

@inproceedings{zhou2022conditional,
  title={Conditional prompt learning for vision-language models},
  author={Zhou, Kaiyang and Yang, Jingkang and Loy, Chen Change and Liu, Ziwei},
  booktitle={Proceedings of the IEEE/CVF conference on computer vision and pattern recognition},
  pages={16816--16825},
  year={2022}
}

@inproceedings{ma2024improved,
  title={Improved self-training for test-time adaptation},
  author={Ma, Jing},
  booktitle={Proceedings of the IEEE/CVF Conference on Computer Vision and Pattern Recognition},
  pages={23701--23710},
  year={2024}
}

@article{jing2019neural,
  title={Neural style transfer: A review},
  author={Jing, Yongcheng and Yang, Yezhou and Feng, Zunlei and Ye, Jingwen and Yu, Yizhou and Song, Mingli},
  journal={IEEE transactions on visualization and computer graphics},
  volume={26},
  number={11},
  pages={3365--3385},
  year={2019},
  publisher={IEEE}
}

@article{gong2022note,
  title={Note: Robust continual test-time adaptation against temporal correlation},
  author={Gong, Taesik and Jeong, Jongheon and Kim, Taewon and Kim, Yewon and Shin, Jinwoo and Lee, Sung-Ju},
  journal={Advances in Neural Information Processing Systems},
  volume={35},
  pages={27253--27266},
  year={2022}
}

@article{llava,
  title={Visual instruction tuning},
  author={Liu, Haotian and Li, Chunyuan and Wu, Qingyang and Lee, Yong Jae},
  journal={Advances in neural information processing systems},
  volume={36},
  pages={34892--34916},
  year={2023}
}

\end{document}